%% file: acl_latex.tex
\pdfoutput=1
\documentclass[11pt]{article}

\usepackage[final]{acl}

\usepackage{times}
\usepackage{latexsym}

\usepackage{multirow}
\usepackage[normalem]{ulem}
\useunder{\uline}{\ul}{}
\usepackage{tabularx}

\usepackage[T1]{fontenc}
\usepackage[utf8]{inputenc}

\usepackage{microtype}

\usepackage{inconsolata}

\usepackage{amsmath}
\usepackage{float}

\usepackage{graphicx}
\usepackage{subcaption}

\usepackage{algorithm}
\usepackage{algpseudocode}
\usepackage{setspace}
\usepackage{caption}
\usepackage{float}

\usepackage{hyperref}
\title{Cleanse: Uncertainty Estimation Approach Using Clustering-based Semantic Consistency in LLMs}

\author{Minsuh Joo ~~~~ Hyunsoo Cho \\
Ewha Womans University \\
\texttt{\{judyjoo21, chohyunsoo\}@ewha.ac.kr}
}

\begin{document}
\maketitle 

    % Main sections
    \input{sections/00_abs}
    \input{sections/01_intro}

    \input{sections/02_related}

\input{sections/03_method}
    \input{sections/04_exp}

\input{sections/05_con}

    \clearpage
    % References
    % \bibliography{reference}
    \bibliographystyle{acl_natbib}

\input{acl_latex.bbl}
    % Appendix
    \input{sections/99_app}

\end{document}

%% file: sections/00_abs.tex
\begin{abstract}
    Despite the outstanding performance of large language models (LLMs) across various NLP tasks, hallucinations in LLMs--where LLMs generate inaccurate responses--remains as a critical problem as it can be directly connected to a crisis of building safe and reliable LLMs. Uncertainty estimation is primarily used to measure hallucination levels in LLM responses so that correct and incorrect answers can be distinguished clearly. This study proposes an effective uncertainty estimation approach, \textbf{Cl}ust\textbf{e}ring-based sem\textbf{an}tic con\textbf{s}ist\textbf{e}ncy (\textbf{Cleanse}). Cleanse quantifies the uncertainty with the proportion of the intra-cluster consistency in the total consistency between LLM hidden embeddings which contain adequate semantic information of generations, by employing clustering. The effectiveness of Cleanse for detecting hallucination is validated using four off-the-shelf models, LLaMA-7B, LLaMA-13B, LLaMA2-7B and Mistral-7B and two question-answering benchmarks, SQuAD and CoQA. 

\end{abstract}

%% file: sections/01_intro.tex
\section{Introduction}

% LLM의 최근 발전과 다재다능함.
Recent advances in LLMs have dramatically enhanced their performance across a wide spectrum of downstream tasks, from translation and summarization to question answering (QA) and dialogue generation. These models now produce fluent, contextually aware outputs that often rival human-like language generation.
% LLM 환각 소개
Despite these remarkable capabilities, a persistent and critical limitation remains: LLMs frequently generate hallucinated outputs—responses that may appear coherent and plausible but are in fact factually incorrect or unsupported by any underlying knowledge \cite{ji2023survey, huang2025survey}. These hallucinations are particularly insidious because they are difficult for users, especially non-experts, to detect, potentially leading to serious consequences in high-stakes applications.
% LLM 환각은 QA처럼 정답의 정확성이 요구되는 task에서 더 심각.
This challenge becomes especially pronounced in QA tasks, where correctness can be objectively verified. Unlike open-ended tasks such as dialogue or summarization—where diverse outputs can still be acceptable—QA typically demands precise and verifiable answers \cite{zhang2023siren}. As a result, even minor hallucinations can significantly degrade task accuracy. When hallucinated outputs are presented in such contexts, they can mislead users, erode trust in AI systems, and compromise the reliability of LLM-based applications \cite{zhang2023siren}. Ensuring the factual consistency of outputs is thus not only a technical concern but also a crucial factor for user safety and system credibility.

% 환각 해결 방법론들 소개
To address these challenges, researchers have proposed a variety of solutions, including dataset refinement, retrieval-augmented generation (RAG), and uncertainty estimation. Each of these approaches targets hallucination from a different angle, offering complementary benefits.
% 다양한 방법론들 소개
One approach is dataset refinement, which involves carefully reviewing and editing training data to improve model accuracy. While this can help reduce errors, it is also highly labor-intensive and difficult to scale. 
Another strategy is retrieval-augmented generation (RAG). By retrieving external knowledge during the generation process, RAG can provide more factually grounded answers. 
However, this approach requires building more complex and potentially fragile pipelines that demand significant computational resources \cite{ji2023survey, es2024ragas}.
% Uncertainty estimation 소개 및 장점
In contrast, uncertainty estimation offers a lightweight and scalable alternative by assessing the model’s confidence in its own outputs. 
Importantly, this method does not require additional external knowledge sources or significant changes to the model architecture. 
Instead, it provides users with interpretable confidence signals that can help identify potentially unreliable responses \cite{lin2022teaching}.
In QA and related tasks, these confidence metrics can serve as a critical line of defense against the unintended consequences of hallucination.

% NLP에서의 uncertainty estimation 소개
Within natural language processing (NLP), uncertainty estimation is typically grounded in the assumption that models are more consistent when confident. 
That is, when a model is certain about its answer, repeated generations will tend to converge; conversely, a lack of confidence often results in high output variability.
% NLP에서 Uncertainty 측정 방법
To assess uncertainty in generated outputs, researchers have proposed methods that operate at various linguistic levels—token and sentence—each providing distinct advantages based on the desired granularity of analysis. 
Token-level metrics such as Perplexity \cite{ren2023outofdistributiondetectionselectivegeneration}, LN-Entropy \cite{malinin2020uncertainty}, and Lexical Similarity \cite{lin2022towards} are well-suited for capturing fine-grained variations within specific output spans, particularly within answer segments of a sentence. 
In contrast, \citet{rabinovich2023predicting} evaluates uncertainty at the sentence-level, making it more appropriate for assessing broader linguistic properties such as overall semantic sentiment. While analyses at both token and sentence levels offer valuable insights, semantic aspect of natural language is more significant when deciding whether two texts with different form are equivalent or not. This is because the inherent variability of natural language data leads to semantic equivalence, where diverse expressions can convey the same meaning \cite{kuhn2023semantic}. Even if two texts use different tokens and syntactic structures, it is reasonable to consider them consistent as long as their semantics are the same. 
% 하지만 기존 cosine score에도 문제가 있음
However, sentence-level similarity measures are not without limitations. \citet{rabinovich2023predicting} calculates all pairwise similarities and they take the average of these similarities equally. It might lead to an incorrect result that a few highly similar sentence pairs disproportionately influence the overall uncertainty score. This can mask the presence of semantically divergent outputs and falsely suggest high consistency.

% 여기에서, 우리 논문 소개
To overcome these challenges and make metric more precise, we introduce \textbf{Cl}ust\textbf{e}ring-based Sem\textbf{an}tic Con\textbf{s}ist\textbf{e}ncy (Cleanse)—a novel sentence-level uncertainty estimation technique designed to more reliably detect hallucinations in generative models.
% High-level  idea
Cleanse leverages bi-directional natural language inference (NLI) to determine whether pairs of generated responses entail one another, forming semantically equivalent clusters with greater precision and excluding any connections that do not meet entailment criteria.
% 좀 더 자세한 설명
We then measure the internal connectivity of these clusters by computing the cosine similarity of their hidden representations as a proxy for semantic consistency, while the distances between clusters provide signals for semantic divergence.  
In other words, dense intra-cluster links indicate semantic agreement, while high inter-cluster links suggest uncertainty. Thus, we estimate uncertainty by leveraging the similarity between embeddings within the same clusters as the degree of consistency. 
% 장점 재어필
By prioritizing these semantically meaningful clusters—rather than relying on simple average similarity—Cleanse offers more calibrated and trustworthy uncertainty estimates. 
% 방법론 좋다 요약
Experiments on QA benchmarks further demonstrate that Cleanse consistently outperforms existing token- and sentence-level methods in detecting hallucinations. We also verify that our key concept, which considers the degree of inter-cluster links (i.e., inter-cluster similarity) as penalty and degree of intra-cluster links (i.e., intra-cluster similarity) as consistency between outputs, contributes to improving hallucination detection performance and the robustness of Cleanse. 

\begin{figure*}[t]
      \centering
      \includegraphics[width=\textwidth]{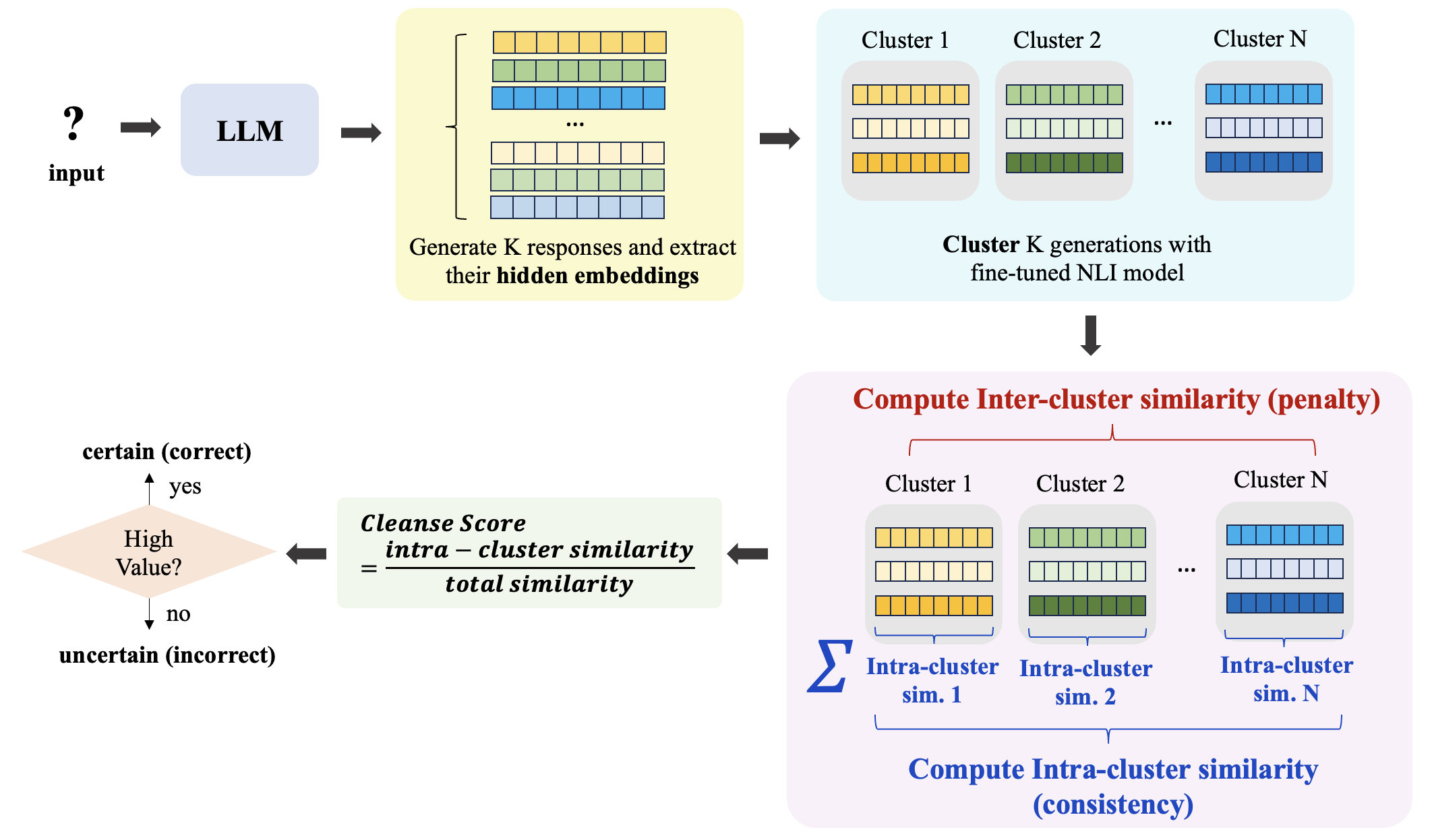}
      \caption {Illustration of Cleanse pipeline.}
      \label{sec:fig1}
    \end{figure*}

%% file: sections/02_related.tex
\section{Related Work}
    There are several related works about uncertainty estimation with various perspectives. The researchers fine-tune the model to ensure that the estimated uncertainty aligns with the actual uncertainty \cite{lin2022teaching}. Application of perturbation module and aggregation module to calibrate uncertainty is an effective setting as well. \cite{gao2024spuq}. Semantic entropy is the entropy across groups clustered by semantically-equivalent outputs \cite{kuhn2023semantic}. Shifting Attention to Relevance (SAR) shifts weights from semantically-irrelevant tokens to semantically-relevant tokens so that probability of relevant tokens contributes to uncertainty quantification more significantly \cite{duan2023shifting}. Recently, there are some approaches using LLM's internal states. The researchers propose a framework named INSIDE, which exploits the eigenvalues of responses' covariance matrix to measure the semantic consistency in the dense embedding space \cite{chen2024inside}. Internal states can be considered as the input of the uncertainty estimator model so that the model classifies whether the response is hallucinated or not \cite{ji2024llm}.

%% file: sections/03_method.tex
\section{Method}

% Comment: 방법론에서 첫 문단은 (1) 우리 방법론에 대한 장점 어필 (2)각 subsection에서 설명할 component들이 어떠한 식으로 동작할지에 대한 high level 설명을 하면 됩니다.(디테일한 내용은 최대한 생략)

% 이후 서브섹션에서 방법론에 들어가는 각 component들을 상세 설명을 하면 더 좋은 논문이 될 것 같아요.지금도 잘 썼지만 좀 더 위 부분을 더 고려해서 작성하면 더 좋아 보입니다.

% 간단히 써봤어요
% Cleanse introduces a novel approach to uncertainty estimation by evaluating how consistently a LLM responds to identical inputs, not just at the surface level but through the semantic structure of its hidden representations.
% Specifically, we first generate multiple responses from a single input and extract the corresponding hidden representations from LLM.
% Then, these representations are grouped leveraging clustering based on bi-directional entailment.
% Finally, the method quantifies uncertainty by analyzing similarity patterns within and across these clusters—greater dispersion across groups indicates higher uncertainty.
% The following subsections describe each stage of the process in detail.

Cleanse estimates the uncertainty by quantifying the intra-cluster consistency between generations, leveraging semantics of responses by employing sentence-level embeddings and bi-directional clustering. 
First, we generate multiple outputs and extract their hidden embeddings from the model. Then, we cluster those outputs based on their semantic equivalency. 
Finally, to assess uncertainty, we compute similarities within and across these clusters respectively and calculate Cleanse Score.
Specifically, we demonstrate the hidden embeddings we use in Section~\ref{sec:3.1}, the clustering technique we use in Section~\ref{sec:3.2}, and how to compute Cleanse score in Section~\ref{sec:3.3}.

\subsection{Hidden embeddings}\label{sec:3.1}

We use the last token embedding in the middle layer of LLM as the output's hidden embedding, as prior work suggests it may capture semantic information effectively \cite{azaria2023internal}. Here, considering a single hidden embedding as a \textit{d}-dimensional vector embedding, we measure the consistency between these hidden embeddings using cosine similarity. 

\subsection{Clustering techniques}\label{sec:3.2}

We apply the concepts used in clustering validation by adapting them to be suitable for our study, which aims for the better and clearer quantification. In general, the main goal of clustering is to maximize the inter-cluster distances and minimize the intra-cluster distances \cite{ansari2015quantitative} and these two criteria are utilized in the clustering validation techniques such as Dunn's Index \cite{ansari2015quantitative}. Dunn's Index is defined as the ratio between the minimum distance across different clusters and the maximum distance within the same cluster, where a value closer to 1 indicates better clustering performance. Here, we could shift the perspective from distance to similarity by taking the inverse of the distance \cite{ansari2015quantitative}. In the perspective of similarity, better clustering corresponds to high intra-cluster similarity and low inter-cluster similarity. When we view it from a consistency perspective rather than clustering validation, it provides an intuitive insight that high intra-cluster similarity indicates the presence of many embeddings sharing equivalent meanings, while high inter-cluster similarity suggests the presence of embeddings with diverse meanings. We perform clustering on the \textit{K} outputs to utilize these similarity concepts. We will further explain what is done with the clustering results in Section~\ref{sec:3.3}. The thing is that, our study aims to compute these similarities and quantify uncertainty, not to minimize inter-cluster similarity or maximize intra-cluster similarity. We just got an intuition from the concept of the distance defined in the clustering, which can be transformed to similarity. 

% 리뷰 후 수정한 것: clustering 방법론 디테일 추가
% bi-directional entailment 설명 추가
% clustering process가 computationally efficient한 이유 자세히 설명
To ensure that the outputs are clustered based on their semantic information, we use a fine-tuned NLI model that maps the input to a high-dimensional semantic embedding. We utilize the clustering algorithm used in the precedent study \cite{kuhn2023semantic}. Here, we introduce only some main concepts for this algorithm. First main concept is that a pair of outputs is considered \verb|entailment| only when both outputs are entail to each other--i.e., bi-directional entailment--which ensures the two outputs truly share the same meaning. Second, researchers concatenated question and its answer in the form of \verb|<Question+Answer>|, insisting that the content of question helps the clustering model comprehend the input context better. Finally, the algorithm is computationally efficient for two reasons. First, the NLI model is substantially smaller than the main model which generates outputs. While the main model has 7B and 13B parameters, the clustering model we used (i.e., nli-deberta-v3-base) has only 184M parameters, making the clustering process comparatively lightweight. Additionally, the number of comparisons required to determine whether an output should be included in the cluster is reduced due to the transitive characteristic between outputs. This transitivity means that a new output can be added to a certain cluster as long as it has a bi-directional entailment with at least one existing member of that cluster, thereby making the number of comparisons be small. More detailed about the algorithm we refer is shown in Algorithm~\ref{alg:alg1}. 

\begin{center}
\begin{minipage}{0.45\textwidth}
\centering
\begin{algorithm}[H]
\caption{Bi-directional Entailment Algorithm}
\label{alg:alg1}
\small
\setstretch{1.2}
\begin{algorithmic}
\Require context $x$, set of seqs. $\{s^{(2)},\dots,s^{(M)}\}$, NLI classifier $\mathcal{M}$, set of meanings $C=\{\{s^{(1)}\}\}$
\For{$2 \leq m \leq M$}
  \For{$c \in C$}
    \State $s^{(c)} \gets c_0$ \hfill{\footnotesize$\triangleright$ Compare to existing meanings}
    \State \texttt{left} $\gets \mathcal{M}(\text{cat}(x, s^{(c)}, ``\!<\!g/\!>\!", x, s^{(m)}))$
    \State \texttt{right} $\gets \mathcal{M}(\text{cat}(x, s^{(m)}, ``\!<\!g/\!>\!", x, s^{(c)}))$
    \If{\texttt{left} \textbf{and} \texttt{right} are \texttt{entailment}}
      \State $c \gets c \cup \{s^{(m)}\}$ \hfill{\footnotesize$\triangleright$ Add to cluster}
    \EndIf
  \EndFor
  \State $C \gets C \cup \{s^{(m)}\}$ \hfill{\footnotesize$\triangleright$ New cluster}
\EndFor
\State \Return $C$
\end{algorithmic}
\end{algorithm}
\end{minipage}
\end{center}
\vspace{1em}

\begin{figure}[]
  \includegraphics[width=0.4\textwidth]
  {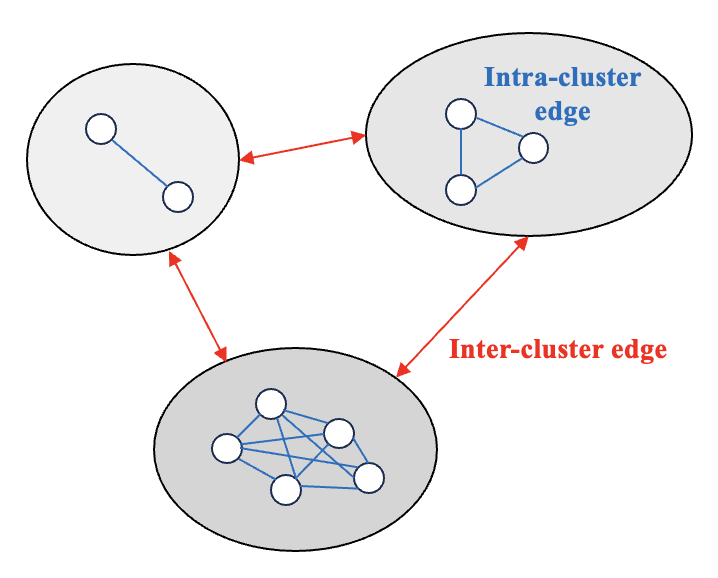}
  \centering
  \caption {Each white circle indicates a single hidden embedding. Edge means the relationship formed between two embeddings. The red edges represent inter-cluster edges, while the blue edges represent intra-cluster edges. Even the red edges are simplified in this illustration, they represent all possible combinations of embeddings in the different clusters. There are given weights to all edges and each of the weight is the computed cosine similarity between two embeddings.}
  \label{sec:fig2}
\end{figure}

\subsection{Cleanse Score}\label{sec:3.3}
% clustering clarify which is the actual weights and which is the penalty weights. 
Here, we define concepts of similarities from Section~\ref{sec:3.2} for clear understanding. Intra-cluster similarity refers the sum of all cosine similarities between embeddings within the same cluster which is computed by Eq~\ref{eq:eq1}. \textit{C} is the number of clusters, $N_k$ is the number of hidden embeddings in the \textit{k}-th cluster, and $\operatorname{cosine(e_i, e_j)}$ is the cosine similarity between \textit{i}-th and \textit{j}-th hidden embeddings. Inter-cluster similarity refers that of all cosine similarities between embeddings across the different clusters. Total similarity is the summation of intra-cluster similarity and inter-cluster similarity which is computed by Eq.~\ref{eq:eq2} where \textit{K} is the number of outputs. Figure~\ref{sec:fig2} clarifies the definition of our terms.

\begin{equation}
    \label{eq:eq1}
    \operatorname{intra\mathchar`-cluster\ sim.} = 
    \sum_{k=1}^{C} 
    \sum_{i=1}^{N_k-1}
    \sum_{j=i+1}^{N_k}
    \operatorname{cosine(e_i, e_j)}
\end{equation}

\begin{equation}
    \label{eq:eq2}
    \operatorname{total\ sim.} = 
    \sum_{i=1}^{K-1}
    \sum_{j=i+1}^{K}
    \operatorname{cosine(e_i, e_j)}
\end{equation}

% 리뷰어 지적: refined consistency의 정의가 명확하지 않다
% 이 부분에 그 정의가 명시되어 있다고 생각하는데, 읽어보시고 어떻게 고치면 좋을 지 알려주시면 감사하겠습니다
By clustering the outputs based on their semantic equivalency, we can identify how many clusters are formed, which in turn indicates how much semantically-inconsistent the outputs are. If there are many clusters, outputs have low consistency (i.e., high uncertainty). In this case, most edges are inter-cluster edges, meaning the inter-cluster similarity is greater than intra-cluster similarity and it leads to low proportion of intra-cluster similarity in the total similarity. In contrast, if the number of clusters is small, outputs have high consistency (i.e., low uncertainty) where most edges are intra-cluster edges. It would lead to high proportion of intra-cluster similarity in the total similarity. Based on this intuition, we measure intra-cluster similarity as the degree of consistency which contributes to the high consistency because they are the similarities between embeddings which are semantically equivalent. Inter-cluster similarity is considered as the penalty for the consistency between outputs as high inter-cluster similarity indicates that there are many outputs belonging to different clusters with divergent meanings. We do clustering in Section~\ref{sec:3.2} in order to map outputs to semantic space and compute inter-cluster similarity and intra-cluster similarity separately. 

\begin{figure}[H]
  \includegraphics[width=0.5\textwidth]{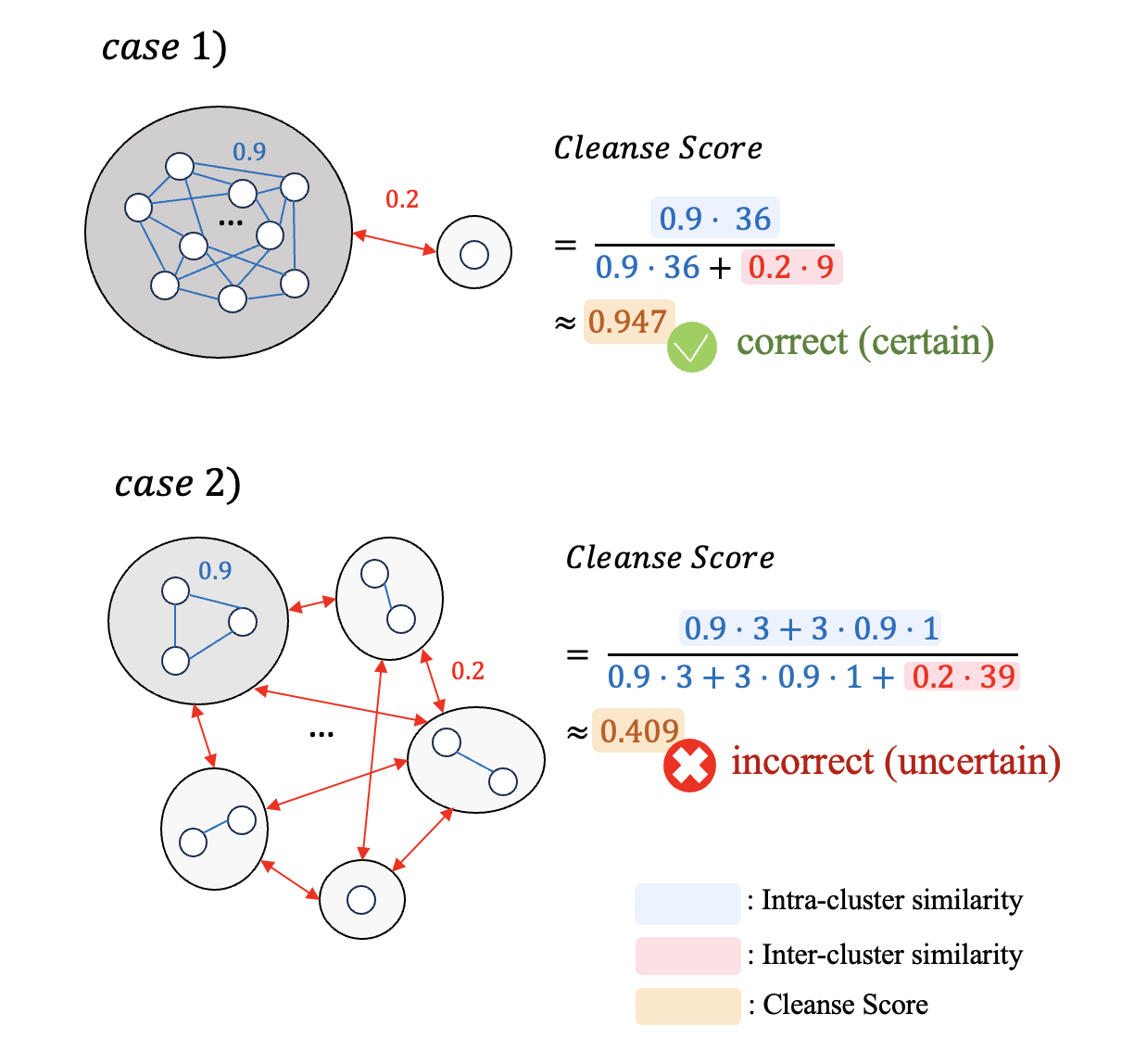}
  \centering
  \caption {Case 1 has a small number of clusters, resulting a high proportion of the intra-cluster similarity in the total similarity. This case will be classified as correct as Cleanse Score is sufficiently high as 0.947, indicating low uncertainty. However, in Case 2, the proportion of the intra-cluster similarity in the total similarity is low at 0.409, so this case will be determined to be incorrect with high uncertainty.}
  \label{sec:fig3}
\end{figure}

We subtract the proportion of inter-cluster similarity in the total similarity from 1, which is the total proportion. Eq.~\ref{eq:eq3} represents how to compute Cleanse Score using two types of similarities. There are two cases in Figure~\ref{sec:fig3}, which shows how does Cleanse Score work effectively and clearly in quantifying consistency. 
\begin{equation}
    \label{eq:eq3}
    \begin{aligned}
        \operatorname{Cleanse\ Score} &= 
        1 - \frac{\operatorname{inter-cluster\ sim.}}{\operatorname{total\ sim.}} \\
        &\vspace{0.1cm} \\ % 줄 간격 추가
        &= \frac{\operatorname{intra-cluster\ sim.}}{\operatorname{total\ sim.}}
    \end{aligned}
\end{equation}

%% file: sections/04_exp.tex
\section{Experiment}

\subsection{Experimental setups}

\begin{table*}[]
\renewcommand{\arraystretch}{1.2}
\centering
{\fontsize{10pt}{11pt}\selectfont

\begin{tabularx}{\textwidth}{cc|cc|cc|cc|cc}
\hline
\multicolumn{2}{c|}{Model}                                                                                               & \multicolumn{2}{c|}{LLaMA-7B} & \multicolumn{2}{c|}{LLaMA-13B} & \multicolumn{2}{c|}{LLaMA2-7B} & \multicolumn{2}{c}{Mistral-7B} \\ \hline
\multicolumn{2}{c|}{Dataset}                                                                                             & SQuAD         & CoQA          & SQuAD          & CoQA          & SQuAD          & CoQA          & SQuAD          & CoQA          \\ \hline
\multicolumn{1}{c|}{\multirow{2}{*}{\begin{tabular}[c]{@{}c@{}}Perplexity\\ (token-level)\end{tabular}}}         & AUC & 60.2          & 66.1          & 61.4           & 63.6          & 63.8           & 62.2          & 53.3           & 57.3          \\
\multicolumn{1}{c|}{}                                                                                            & PCC   & 19.3          & 27.4          & 21.8           & 27.0          & 25.5           & 24.3          & 13.0           & 21.7          \\ \hline
\multicolumn{1}{c|}{\multirow{2}{*}{\begin{tabular}[c]{@{}c@{}}LN-Entropy\\ (token-level)\end{tabular}}}         & AUC & 72.3          & 71.6          & 74.6           & 70.8          & 74.2           & 70.5          & 59.3           & 61.7          \\
\multicolumn{1}{c|}{}                                                                                            & PCC   & 38.9          & 35.5          & 43.6           & 37.1          & 42.8           & 34.7          & 14.8           & 24.6          \\ \hline
\multicolumn{1}{c|}{\multirow{2}{*}{\begin{tabular}[c]{@{}c@{}}Lexical Similarity\\ (token-level)\end{tabular}}} & AUC & 76.9          & 76.1          & 78.9           & 75.6          & 80.4           & 76.2          & 69.0           & 74.9          \\
\multicolumn{1}{c|}{}                                                                                            & PCC   & 51.2          & 47.7          & 54.4           & 49.1          & 57.4           & 48.6          & 31.4           & 43.2          \\ \hline
\multicolumn{1}{c|}{\multirow{2}{*}{\begin{tabular}[c]{@{}c@{}}Cosine Score\\ (sentence-level)\end{tabular}}}    & AUC & 79.6          & 78.5          & 81.1           & 77.7          & 82.1           & 79.3          & 65.9           & 74.1          \\
\multicolumn{1}{c|}{}                                                                                            & PCC   & 54.7          & \textbf{48.4} & 57.8           & 49.3          & 59.7           & \textbf{50.6} & 29.1           & 41.3          \\ \hline
\multicolumn{1}{c|}{\multirow{2}{*}{\begin{tabular}[c]{@{}c@{}}Cleanse Score\\ (sentence-level)\end{tabular}}}      & AUC & \textbf{81.7} & \textbf{79.4} & \textbf{82.8}  & \textbf{79.6} & \textbf{83.0}  & \textbf{80.1} & \textbf{75.9}  & \textbf{80.2} \\
\multicolumn{1}{c|}{}                                                                                            & PCC   & \textbf{56.4} & 47.6          & \textbf{59.6}  & \textbf{50.7} & \textbf{61.0}  & 49.7          & \textbf{41.6}  & \textbf{47.2} \\ \hline

\end{tabularx}
}
\caption{Hallucination detection performance for four models and two question-answering datasets. AUROC (AUC) and PCC are utilized to evaluate the performance of four baselines and Cleanse Score. We use Rouge-L threshold as 0.7 and deberta-nli-v3-base as a clustering model. Token-level indicates that corresponding metric estimates uncertainty based on token-probability or lexical form of generations. Sentence-level indicates that corresponding metric utilizes sentence-level embedding in computing uncertainty. Bolded values indicate the highest scores.}
\label{tab:table1}
\end{table*}

\paragraph{Datasets.} We use two representative question-answering datasets, SQuAD \cite{rajpurkar2016squad} and CoQA \cite{reddy2019coqa}. SQuAD (20.92) has longer ground truth answer spans than CoQA (13.67) when we compute the average of the length of golden answer for each dataset in our experiment. We follow the prompt setting of SQuAD as presented by \citet{chen2024inside} and that of CoQA as described by \citet{lin2023generating}. 

\paragraph{Models.} We conduct experiments by varying the model in terms of its size, version, and optimized method. We utilize four off-the-shelf models, LLaMA-7B \cite{llama}, LLaMA-13B \cite{llama}, LLaMA2-7B \cite{llama2}, and Mistral-7B \cite{jiang2023mistral7b}.

\paragraph{Baselines.} We compare the performance of Cleanse Score to four baeslines. \textbf{Perplexity} \cite{ren2023outofdistributiondetectionselectivegeneration} measures the total uncertainty for generated sequence using the uncertainty of each token which consists of the sequence. \textbf{Length-normalized entropy (LN-entropy)} \cite{malinin2020uncertainty} is similar to perplexity, but it reduces the bias in quantifying uncertainty by normalizing the joint log-probabilities with its sequence length. \textbf{Lexical similarity} \cite{lin2022towards} is the average similarities between the answers which are measured with Rouge-L \cite{lin2004rouge}. \textbf{Cosine score}, computed as Eq.~\ref{eq:eq4} in our study, serves as a baseline to verify that incorporating inter-cluster similarity as a penalty helps clarify the boundary between certain and uncertain answers, thereby improving uncertainty estimation performance. 

    \begin{equation}
        \label{eq:eq4}
        \text{cosine score} = 
        \frac{2}{K(K-1)}
        \sum_{i=1}^{K-1} 
        \sum_{j=i+1}^{K}
        \operatorname{cosine(e_i, e_j)}
    \end{equation}

\paragraph{Correctness measure.} We use Rouge-L \cite{lin2004rouge} as the correctness measure which determines whether the generation of LLM is correct or not, comparing it with the ground truth answer. We set the threshold as 0.7, which means only generation $\operatorname{s}$ is considered to be correct if $\operatorname{s}$ satisfies $\mathcal{L}(\operatorname{s}, \operatorname{s'}) = {1}_{\operatorname{Rouge-L}(s, s') > 0.7}$ for the ground truth answer $\operatorname{s'}$. We adjust this threshold from 0.5 to 0.9 in our further experiment to demonstrate the general capability of Cleanse Score.

\paragraph{Evaluation measure.} We utilize two evaluation measures to evaluate the uncertainty estimation performance of four baselines and Cleanse Score. We use Area Under the Receiver Operating Characteristic Curve (AUROC) and Pearson Correlation Coefficient (PCC). AUROC is a performance metric for binary classifiers, allowing it to assess whether an uncertainty estimation metric effectively distinguishes between correct and incorrect generations. PCC measures the correlation between the Rouge-L score and the consistency level computed by each metric. Higher AUROC and PCC indicate better performance.

\subsection{Main results}

\paragraph{Effectiveness of Cleanse.} As shown in Table~\ref{tab:table1}, Cleanse Score outperforms all four baselines across LLaMA models and Mistral-7B on the SQuAD and CoQA datasets when evaluated using AUROC and PCC. Cleanse Score consistently achieves the highest AUROC, with a particularly large margin in the Mistral-7B settings. In the Mistral-7B model, Cleanse Score surpasses lexical similarity--the second highest performing baseline in Mistral-7B--by 6.9\% in SQuAD and 5.3\% in CoQA. There is a tendency that the performance of Cleanse Score improves in LLaMA-13B and LLaMA2-7B than LLaMA-7B and Mistral-7B. 

On average, cosine score and Cleanse Score, which both leverage sentence-level embeddings, show better performance than the baselines based on token-probability or lexical similarity. This result supports our discussion in the previous section, demonstrating that prioritizing semantic aspect over lexical aspect is a reasonable approach in determining consistency between texts.
% reulst 표 삽입 (table로 형식 고치기)

Additionally, in Table~\ref{tab:table1}, Cleanse Score outperforms cosine score in all cases when evaluated with AUROC and in most cases when evaluated with PCC. Through this result, we demonstrate that our core intuition—clustering multiple outputs and using the inter-cluster similarity as a penalty term—successfully enhances uncertainty detection performance when applied to Cleanse Score. Interpreting intra-cluster similarity and inter-cluster similarity  as the degree of consistency and inconsistency respectively enables us to filter hallucinated cases better than simply by averaging total similarities. 

\paragraph{Advantage of Cleanse: Superior hallucination detection capability even under strict conditions}
% 시각 그래프로 고치기(2열 2행)

\begin{figure*}
  \centering
  \begin{minipage}{0.49\linewidth}
    \subfloat[LLaMA-7B\label{sec:fig4a}]{\includegraphics[width=\linewidth, height=0.55\linewidth, keepaspectratio]{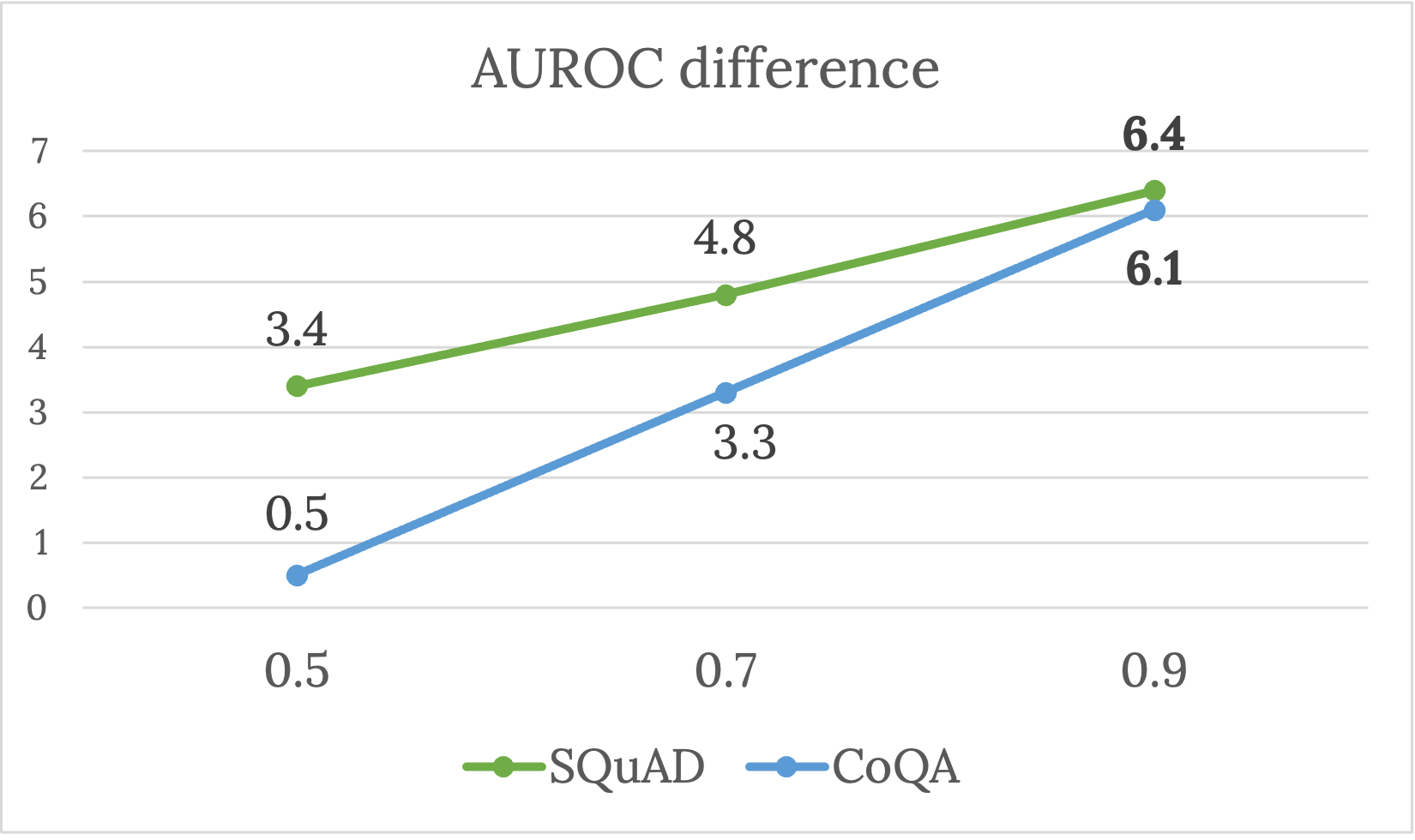}}
  \end{minipage}
  \hfill
  \begin{minipage}{0.49\linewidth}
    \subfloat[LLaMA-13B\label{sec:fig4b}]{\includegraphics[width=\linewidth, height=0.55\linewidth, keepaspectratio]{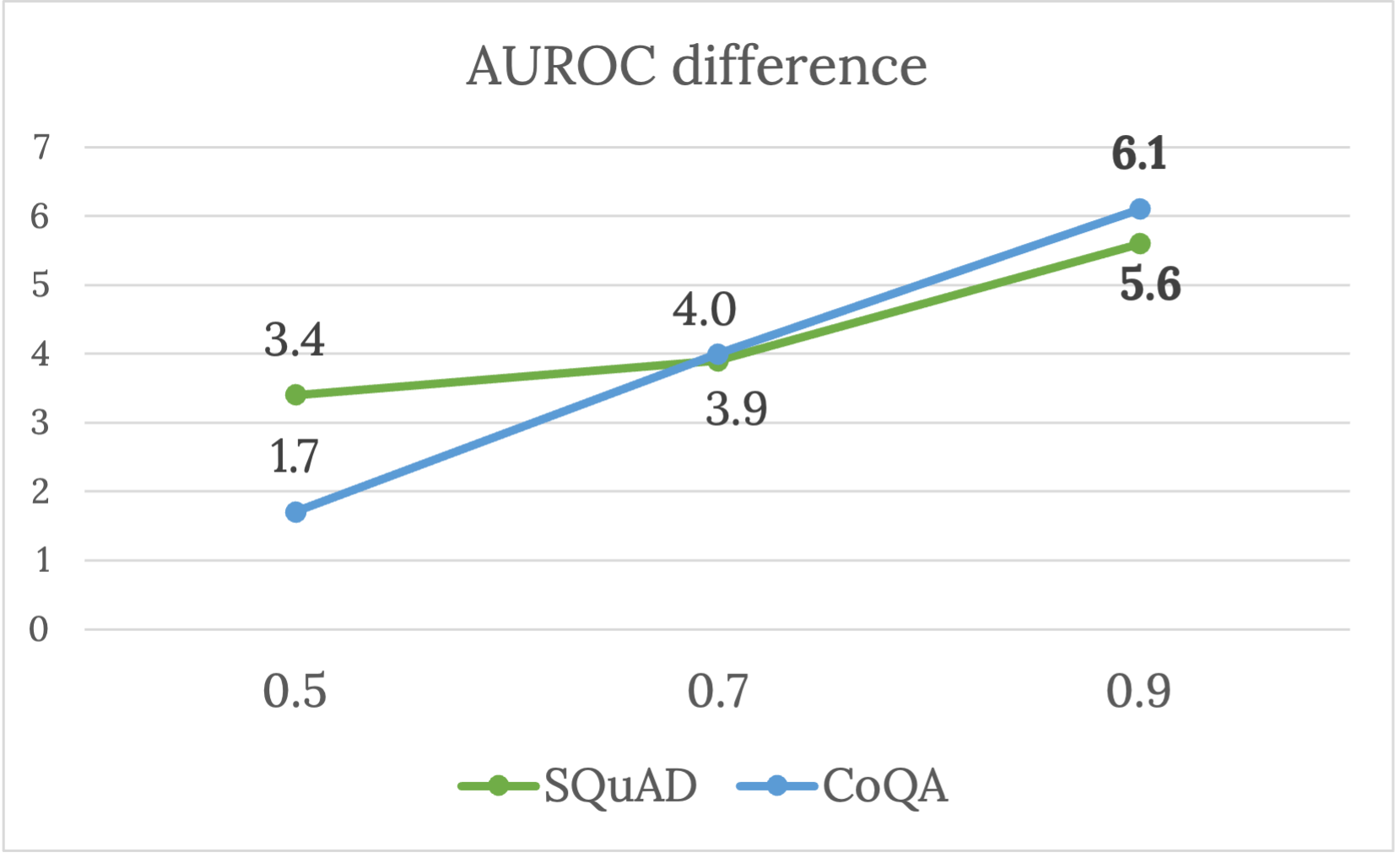}}
  \end{minipage}
  \vfill
  \begin{minipage}{0.49\linewidth}
    \subfloat[LLaMA2-7B\label{sec:fig4c}]{\includegraphics[width=\linewidth, height=0.55\linewidth, keepaspectratio]{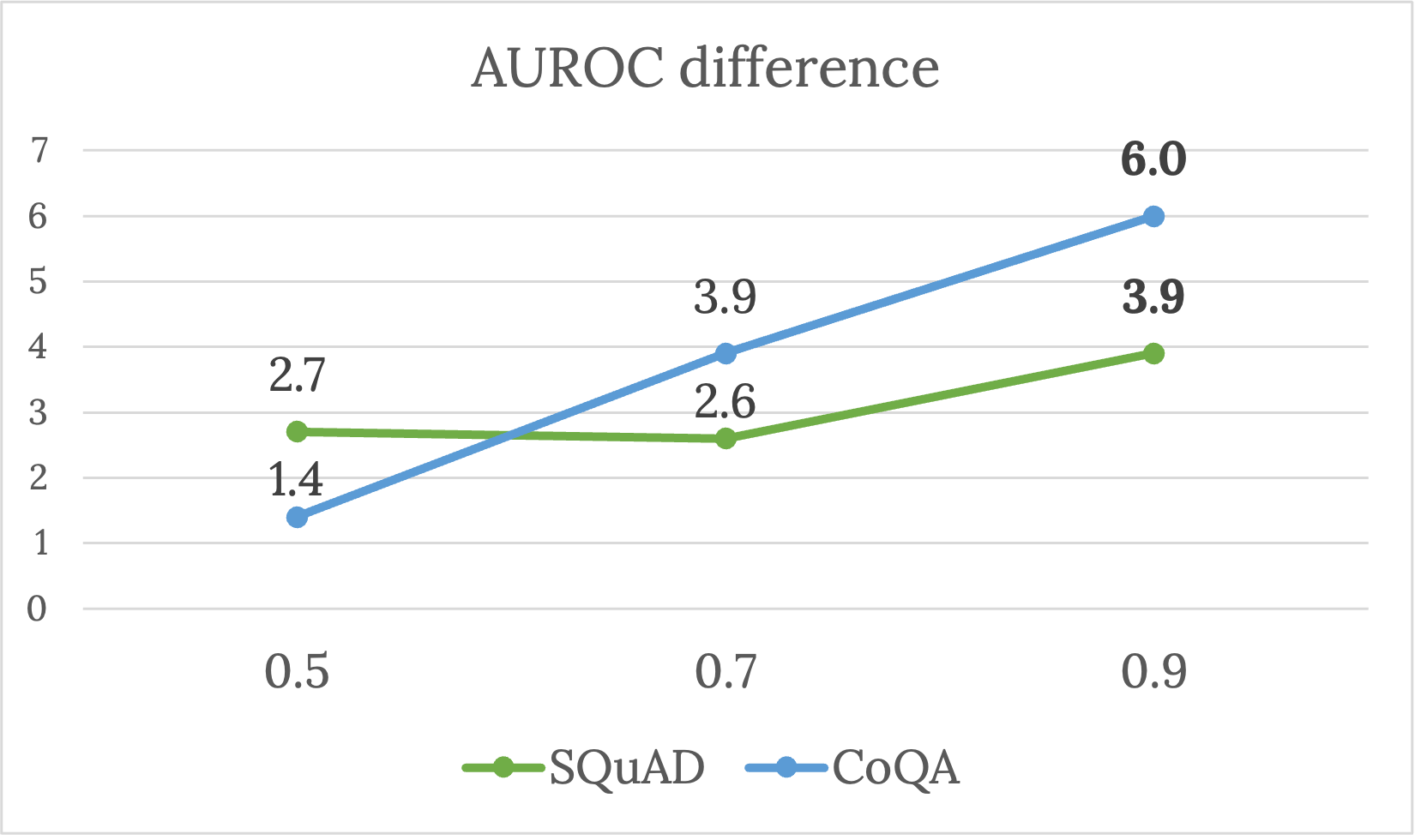}}
  \end{minipage}
  \hfill
  \begin{minipage}{0.49\linewidth}
    \subfloat[Mistral-7B\label{sec:fig4d}]{\includegraphics[width=\linewidth, height=0.55\linewidth, keepaspectratio]{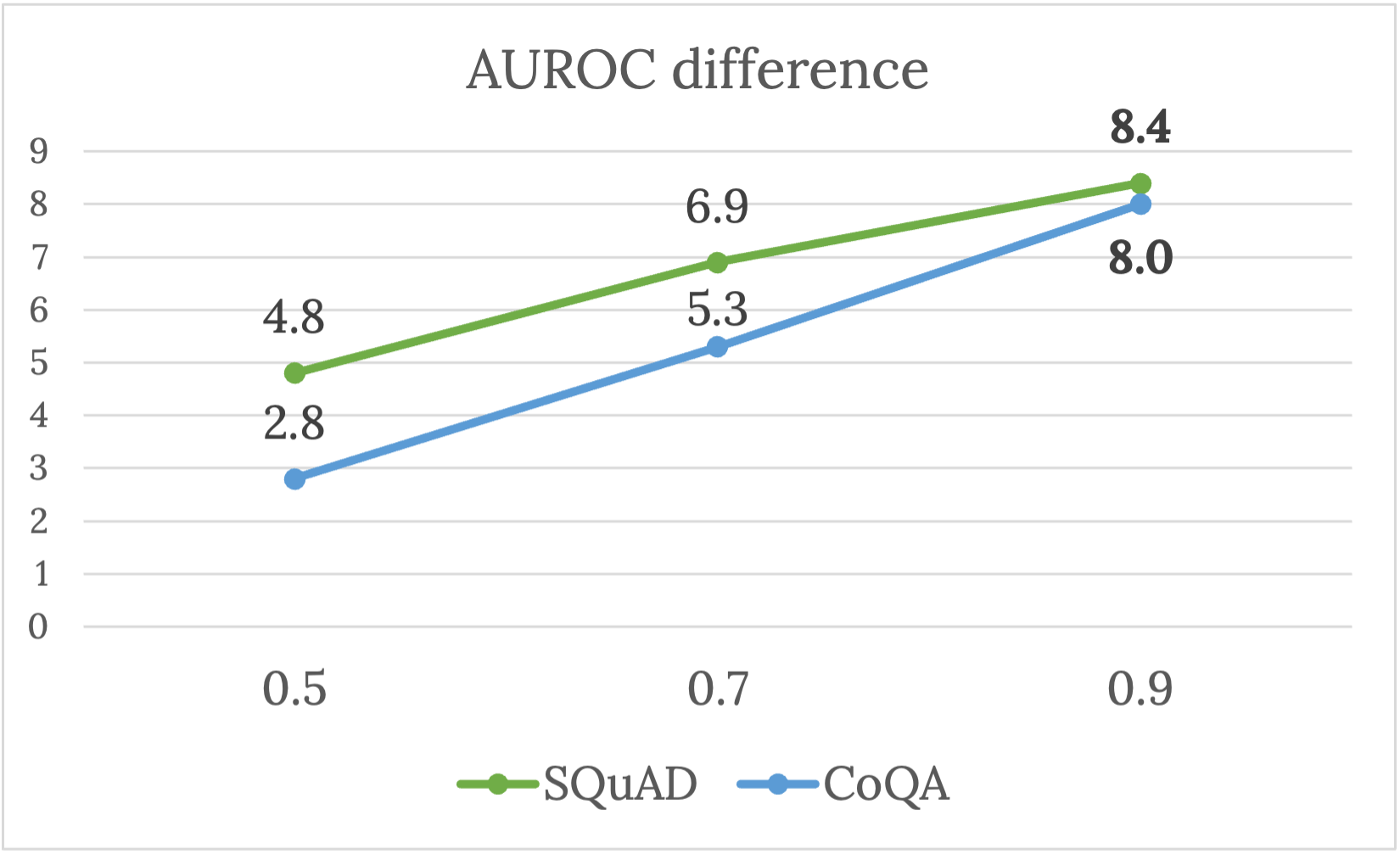}}
  \end{minipage}
  \caption{AUROC difference between Cleanse Score and lexical similarity across four models on two QA datasets, varying the correctness measure threshold between 0.5 to 0.9. The highest values are in bold.}
  \label{sec:fig4}
\end{figure*}

In Figure~\ref{sec:fig4}, we compute the AUROC difference between Cleanse Score and lexical similarity, which achieves the highest performance among token-level approaches. The AUROC differences increase as the threshold of Rouge-L becomes harder, regardless of the model type and dataset. In particular, the differences in LLaMA-7B in Figure~\ref{sec:fig4a} and Mistral-7B in Figure~\ref{sec:fig4d} across both SQuAD/CoQA datasets settings are significant, achieving 6.4\%/6.1\% and 8.4\%/8.0\%. A detailed analysis of the results shown in Table~\ref{tab:table3} in Appendix reveals that, except for the case of Mistral-7B on the SQuAD dataset, the performance of lexical similarity either remains the same or decreases as the Rouge-L threshold increases, whereas the performance of Cleanse Score consistently improves. In the case of Mistral-7B on the SQuAD dataset, the performance of lexical similarity also increases with a higher threshold, but the improvement margin of Cleanse Score is significantly greater that of lexical similarity. Here, increasing the threshold means that the correctness measure becomes more rigorous and aligns more closely with human evaluation. These settings are crucial for certain NLP tasks that require a precise and accurate correctness metric. The results demonstrate that Cleanse Score is robustly applicable in such strict environments such as question-answering and translation tasks.

\begin{figure}[H]
  \includegraphics[width=0.5\textwidth]{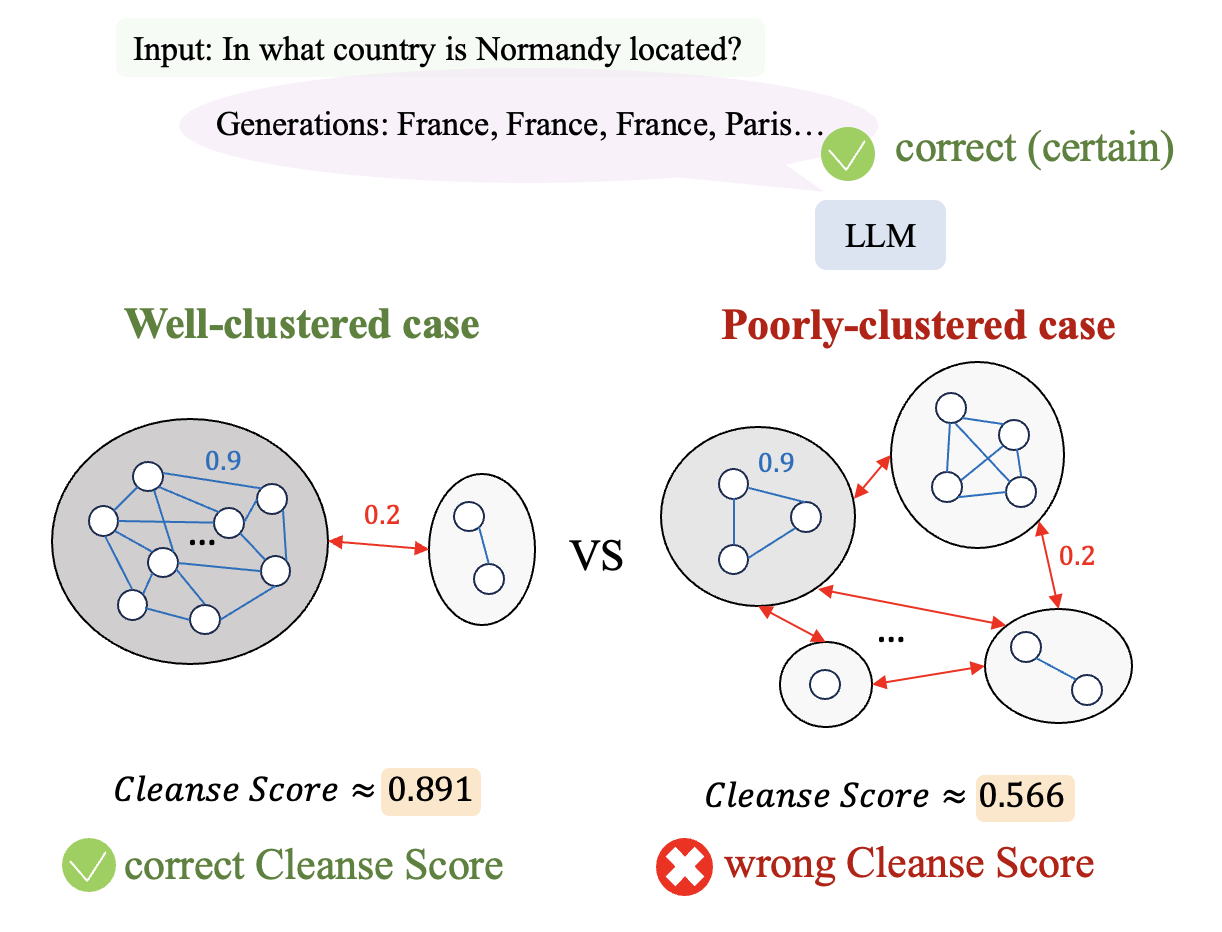}
  \centering
  \caption {The illustration that shows the importance of clustering in our approach. For the same query that the model answers correctly, a well-clustered case results in few clusters, leading to an accurate Cleanse score. In contrast, a poorly-clustered case forms a few scattered clusters which yield an incorrect Cleanse score. This demonstrates that having few clusters for correct answers and a few clusters for wrong answers is advantageous for clearer hallucination detection.}
  \label{sec:fig5}
\end{figure}

\paragraph{Clustering model comparison.} The choice of clustering model is one of the most important factors in our study as shown in Figure~\ref{sec:fig5}. We compare four fine-tuned NLI model, deberta-large-mnli \cite{he2020deberta}, roberta-large-mnli \cite{liu2019roberta}, nli-deberta-v3-base \cite{he2021debertav3} and  nli-deberta-v3-large \cite{he2021debertav3} to find the optimal clustering model. 

We identify the performance of each clustering model in two ways. First, we compare AUROC when each clustering model is applied to Cleanse Score. Table~\ref{tab:table2} shows that AUROC scores of Cleanse Score using nli-deberta-v3-base are slightly better than when using other clustering models.
\begin{table*}[]
\renewcommand{\arraystretch}{1.2}
\centering
{\fontsize{10pt}{11pt}\selectfont
\begin{tabularx}{\textwidth}{cc|c|c|c|c}
\hline
\multicolumn{2}{c|}{Clustering Model}                    & deberta-large-mnli & roberta-large-mnli & nli-deberta-v3-base & nli-deberta-v3-large \\ \hline
\multicolumn{1}{c|}{\multirow{2}{*}{LLaMA-7B}}   & SQuAD & 81.3 (2.71)  & 80.7 (2.54)  & \textbf{81.7 (2.78)} & 81.2 (2.63)           \\
\multicolumn{1}{c|}{}                            & CoQA  & 79.0 (2.49)  & 78.5 (2.40)   & \textbf{79.4 (2.55)} & \textbf{79.4} (2.45)  \\ \hline
\multicolumn{1}{c|}{\multirow{2}{*}{LLaMA-13B}}  & SQuAD & 82.5 (2.96)  & 82.3 (2.78)  & \textbf{82.8 (3.03)} & 82.6 (2.88)           \\
\multicolumn{1}{c|}{}                            & CoQA  & 79.3 (2.47)  & 79.0 (2.36)  & \textbf{79.6 (2.53)} & 79.5 (2.51)           \\ \hline
\multicolumn{1}{c|}{\multirow{2}{*}{LLaMA2-7B}}  & SQuAD & 82.7 (2.92)  & 82.2 (2.73)  & \textbf{83.0 (2.99)} & 82.7 (2.86)           \\
\multicolumn{1}{c|}{}                            & CoQA  & 79.7 (2.52)  & 79.4 (2.43)  & 80.1 \textbf{(2.60)} & \textbf{80.2} (2.57)  \\ \hline
\multicolumn{1}{c|}{\multirow{2}{*}{Mistral-7B}} & SQuAD & 75.2 (1.84)  & 74.2 (1.59)  & \textbf{75.9 (1.92)} & 74.9 (1.75)           \\
\multicolumn{1}{c|}{}                            & CoQA  & 80.0 (2.57)  & 79.4 (2.45)  & \textbf{80.2 (2.63)} & 79.8 (2.55)           \\ \hline

\end{tabularx}
}
\caption{The results of the Cleanse Score performance comparison, measured by AUROC and the difference between the average number of clusters of correct and incorrect answers across four distinct clustering techniques when applied to the methodology (the latter is shown in parentheses). We set Rouge-L threshold as 0.7. Bold values are the highest.}
\label{tab:table2}
\end{table*}
Besides this result, inspired by the intuition from \citet{kuhn2023semantic}, we conduct additional comparison using the concept mentioned in Section~\ref{sec:3.3}. In Figure~\ref{sec:fig5}, a clustering model that forms a small number of clusters for correct answers and a large number of clusters for incorrect answers can clarify  between certain and uncertain outputs, leading Cleanse Score to predict correct and incorrect labels better. Based on this idea, the difference in the number of clusters formed in incorrect generations and correct generations can serve as a metric for evaluating the performance of clustering. The larger the difference is, the better the model clusters. We calculate the difference between the average number of clusters for correct and incorrect generations and show them in parentheses in Table~\ref{tab:table2}. The overall differences for nli-deberta-v3-base are the largest, confirming again that using nli-deberta-v3-base as a clustering model outperforms other models.  

%% file: sections/05_con.tex
\section{Conclusion}

Uncertainty estimation is one of the main solutions in detecting hallucination and prevent it from becoming critical problem in constructing reliable and trustworthy LLMs. We propose Cleanse, which clusters the outputs and computes the proportion of the intra-cluster similarity in the total similarity to quantify the consistency. As a result, filtering inter-cluster similarity as the inconsistency term helps to classify certain and uncertain generations effectively so that Cleanse perform better than the other existing approaches. Also, we found that Cleanse works well even under various correctness measure settings, which indicates Cleanse is appropriate to detecting uncertainty in diverse NLP tasks. Additionally, by conducting further experiments, we could identify a clustering model that outperforms than the others, thereby enhancing the performance of Cleanse. 

\section*{Limitations}

This approach is limited to white-box LLM as it requires hidden embedding extracted directly from the model. However, the performance and usefulness of Cleanse is verified through several experiments, other vector embeddings of the outputs could be used instead of hidden embeddings from a model, thereby overcome this limitation. 

%% file: sections/99_app.tex
\clearpage

\onecolumn
\appendix
\section*{Appendix}

\label{app:A}
% \begin{figure*}[ht]
%   \centering
%   \includegraphics[width=\textwidth]{latex/figs/bi-direction.png}
%   \caption {The algorithm of bi-directional clustering we used in our study. We refer this settings from \citet{kuhn2023semantic}.}
%   \label{sec:fig8}
% \end{figure*}

% \begin{algorithm}[H]
% \caption{Bidirectional Entailment Clustering}
% \label{alg:alg1}
% \begin{algorithmic}
% \Require context $x$, set of seqs. $\{s^{(2)},\dots,s^{(M)}\}$, NLI classifier $\mathcal{M}$, set of meanings $C=\{\{s^{(1)}\}\}$
% \For{$2 \leq m \leq M$}
%   \For{$c \in C$}
%     \State $s^{(c)} \gets c_0$ \hfill{$\triangleright$ Compare to already-processed meanings.}
%     \State \texttt{left} $\gets \mathcal{M}(\text{cat}(x, s^{(c)}, ``<g/>", x, s^{(m)}))$ \hfill{$\triangleright$ Does old sequence entail new one?}
%     \State \texttt{right} $\gets \mathcal{M}(\text{cat}(x, s^{(m)}, ``<g/>", x, s^{(c)}))$ \hfill{$\triangleright$ Vice versa?}
%     \If{\texttt{left} is \texttt{entailment} \textbf{and} \texttt{right} is \texttt{entailment} (bi-directional)}
%       \State $c \gets c \cup \{s^{(m)}\}$ \hfill{$\triangleright$ Put into existing class.}
%     \EndIf
%   \EndFor
%   \State $C \gets C \cup \{s^{(m)}\}$ \hfill{$\triangleright$ Semantically distinct, gets own class.}
% \EndFor
% \State \Return $C$
% \end{algorithmic}
% \end{algorithm}

\section{Additional Experiments}

% \label{app:B}
% \usepackage{multirow}
\begin{table}[H]
\renewcommand{\arraystretch}{1.2}
\centering
{\fontsize{10pt}{12pt}\selectfont
\begin{tabular}{cc|cc|cc|cc|cc}
\hline
\multicolumn{2}{c|}{Model}                                                                                & \multicolumn{2}{c|}{LLaMA-7B} & \multicolumn{2}{c|}{LLaMA-13B} & \multicolumn{2}{c|}{LLaMA2-7B} & \multicolumn{2}{c}{Mistral-7B} \\ \hline
\multicolumn{2}{c|}{Dataset}                                                                              & SQuAD          & CoQA         & SQuAD          & CoQA          & SQuAD          & CoQA          & SQuAD          & CoQA          \\ \hline
\multicolumn{1}{c|}{\multirow{3}{*}{\begin{tabular}[c]{@{}c@{}}Lexical \\ Similarity\end{tabular}}} & 0.5 & 76.8           & 76.9         & 79.1           & 77.1          & 80.2           & 77.5          & 67.6           & 74.9          \\
\multicolumn{1}{c|}{}                                                                               & 0.7 & 76.9           & 76.1         & 78.9           & 75.6          & 80.4           & 76.2          & 69.0           & 74.9          \\
\multicolumn{1}{c|}{}                                                                               & 0.9 & 75.7           & 74.9         & 77.1           & 74.5          & 79.8           & 74.8          & 70.7           & 73.6          \\ \hline
\multicolumn{1}{c|}{\multirow{3}{*}{Cleanse Score}}                                                    & 0.5 & 80.2           & 77.4         & 82.5           & 78.8          & 82.9           & 78.9          & 72.4           & 77.7          \\
\multicolumn{1}{c|}{}                                                                               & 0.7 & 81.7           & 79.4         & 82.8           & 79.6          & 83.0           & 80.1          & 75.9           & 80.2          \\
\multicolumn{1}{c|}{}                                                                               & 0.9 & 82.1           & 81.0         & 82.7           & 80.6          & 83.7           & 80.8          & 79.1           & 81.6          \\ \hline
\end{tabular}
}
\caption{Pattern of AUROC performance changes in lexical similarity and Cleanse Score as Rouge-L threshold varies across 0.5, 0.7, and 0.9. We use deberta-nli-v3-base for clustering model.}
\label{tab:table3}
\end{table}